\documentclass[letterpaper, 10 pt, conference]{ieeeconf}

\IEEEoverridecommandlockouts                              % This command is only needed if % you want to use the \thanks command

\usepackage{graphicx} % for pdf, bitmapped graphics files
\usepackage{hyperref}
\usepackage{mathptmx} % assumes new font selection scheme installed
\usepackage{amsmath} % assumes amsmath package installed
\usepackage{amssymb}  % assumes amsmath package installed
\usepackage{eucal}
\usepackage{algorithm}
\usepackage{algpseudocode}
\usepackage{booktabs}       % professional-quality tables
\usepackage{caption}

\title{\LARGE \bf
Dynamic Obstacle Avoidance through Uncertainty-Based Adaptive Planning with Diffusion
}

\usepackage[utf8]{inputenc}

\author{{Vineet Punyamoorty*$^{1}$, Pascal Jutras-Dubé*$^{1}$, Ruqi Zhang$^{1}$, Vaneet Aggarwal$^{1}$, Damon Conover$^{2}$, Aniket Bera$^{1}$}\\
{\textit{$^1$Purdue University, USA  $^2$DEVCOM Army Research Laboratory, USA}}\\
{\texttt{\{vpunyamo, pjutrasd, ruqiz, vaneet, aniketbera\}@purdue.edu},}\\
{\texttt{damon.m.conover.civ@army.mil}}
 \thanks{* denotes equal contribution}
}

\hypersetup{
colorlinks = true,
urlcolor = blue,
linkcolor = blue,
pdftitle = {ICRA 2025},
pdfauthor = {Vineet Punyamoorty},
pdfstartview={XYZ null null 1.00}
}

\begin{document}
\maketitle
\thispagestyle{empty}
\begin{abstract}
By framing reinforcement learning as a sequence modeling problem, recent work has enabled the use of generative models, such as diffusion models, for planning.
While these models are effective in predicting long-horizon state trajectories in deterministic environments, they face challenges in dynamic settings with moving obstacles. 
Effective collision avoidance demands continuous monitoring and adaptive decision-making. 
While replanning at every timestep could ensure safety, it introduces substantial computational overhead due to the repetitive prediction of overlapping state sequences---a process that is particularly costly with diffusion models, known for their intensive iterative sampling procedure.
We propose an adaptive generative planning approach that dynamically adjusts replanning frequency based on the uncertainty of action predictions.
Our method minimizes the need for frequent, computationally expensive, and redundant replanning while maintaining robust collision avoidance performance. In experiments, we obtain a 13.5\% increase in the mean trajectory length and 12.7\% increase in mean reward over long-horizon planning, indicating a reduction in collision rates, and improved ability to navigate the environment safely.
\end{abstract}

\section{Introduction}\label{sec:introduction}
Diffusion models have recently emerged as a promising approach to planning, demonstrating superior performance across a wide range of domains  \cite{sohl2015thermo, song2019smld, ho_2020_denoising, song2021sde}. Given a dataset of reward-labeled sub-optimal trajectories, diffusion models are capable of stitching them together to generate reward-maximizing optimal trajectories. Unlike single-step autoregressive models, diffusion models enable planning over long horizons without suffering from compounding errors. While such long-horizon planning is beneficial in static environments, it is not well-suited for dynamically changing environments often encountered in reality.

Dynamically varying, realistic environments with a large number of moving obstacles pose significant challenges. Long-horizon planning becomes inadequate in such settings, as rapidly moving obstacles can quickly render the long-term plan obsolete, increasing the likelihood of collisions. Conversely, re-planning at each time step incurs prohibitively high computational costs. In this work, we address this problem through uncertainty-based adaptive diffusion planning for collision avoidance in dynamic environments.

Collision avoidance is a central challenge in planning, control, and robotics, critical for autonomous systems operating in environments with dynamic obstacles. Efficient and reliable mechanisms are essential for self-driving cars navigating unpredictable traffic with vehicles and pedestrians, and industrial robots are adjusting swiftly to changes to ensure safety and delivery drones, avoiding obstacles like birds or other drones. In military applications, autonomous ground vehicles and unmanned aerial systems must navigate complex environments with moving threats, while autonomous underwater vehicles for mine detection and surveillance need to avoid dynamic obstacles to ensure mission success and asset safety.

Traditionally, methods like obstacle potential fields and rule-based approaches, such as the Dynamic Window Approach (DWA) \cite{fox_1997_dynamic} and Timed Elastic Band (TEB) \cite{rosmann_2015_timed}, have been employed for collision avoidance. However, these methods are primarily designed for static obstacles and struggle in dynamic environments. Advancements in Reinforcement Learning (RL) have shown promise in handling dynamic environments by enabling agents to learn from interactions. In this context, deep RL-based algorithms, such as CADRL \cite{chen_2017_decentralized, everett_2021_collision} and MRCA \cite{long_2018_towards}, have demonstrated progress in collision avoidance. More recently, offline RL has emerged as a superior alternative to conventional RL, especially in scenarios where direct interaction with the environment could be dangerous. Offline RL allows agents to learn optimal policies from pre-collected datasets without further exploration. In this context, deep generative models have been effectively applied to sequential decision-making, treating it as a long-sequence modeling task \cite{chen_2021_decision, janner_2022_planning, ajay_2023_conditional}. In real-world, dynamically changing environments with moving obstacles, such long-horizon planning can severely increase the risk of collisions. However, collision avoidance within deep generative model-based offline RL has yet to be explored.

In this work, we introduce a novel solution to collision avoidance by combining diffusion models with adaptive planning strategies, offering a way to both predict and avoid collisions with minimal computational cost. By leveraging uncertainty estimates obtained from a deep ensemble inverse dynamics action model, we demonstrate the ability to dynamically adjust the planning horizon, utilizing computational resources only when necessary. This approach reduces unnecessary re-planning and enhances safety in environments with moving obstacles. 

Our work has the following key contributions:
\begin{itemize}
    \item We propose a novel approach to collision avoidance in dynamically changing environments through adaptive re-planning based on uncertainty estimates obtained from a deep ensemble action dynamics model. 
    \item Our approach provides a tunable and flexible trade-off between long-horizon planning (with high collision risk in a dynamic environment) and re-planning at every step (which is computationally expensive and redundant), through a single tuning parameter.
    \item We demonstrate the effectiveness of our approach through experiments on a dynamic environment involving fast-moving obstacles, \texttt{highway-env} \cite{highway-env}. 
\end{itemize}

In our experiments, we obtain a 13.5\% increase in the mean trajectory length, which indicates a reduction in collision rates, as
longer trajectories suggest that the agent successfully avoids collisions.

\begin{figure*}
    \centering
    \includegraphics[width=0.85\textwidth]{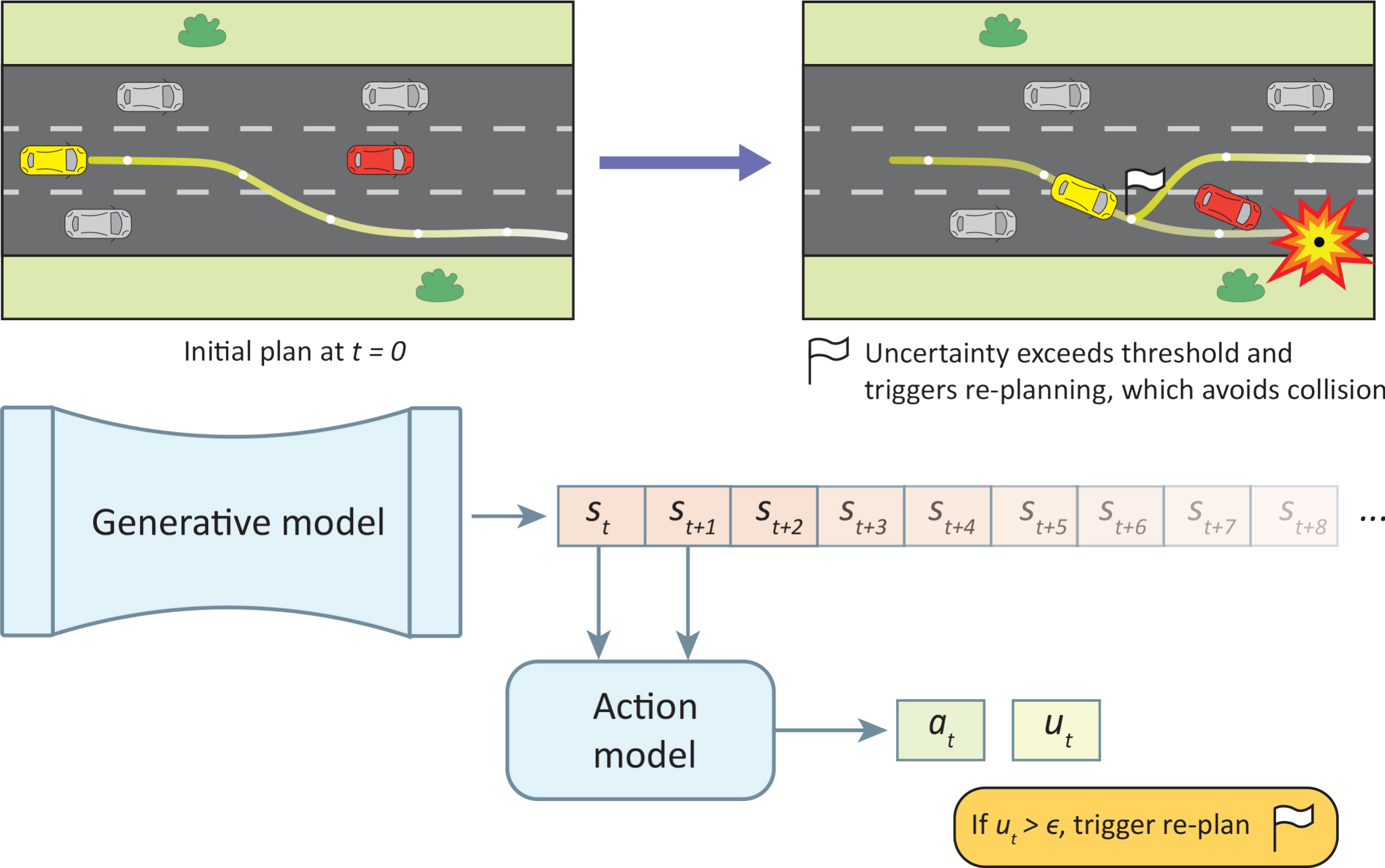}
    \caption{Diffusion model generates an initial long-horizon trajectory based on the past history of states. However, in a dynamically changing environment, moving obstacles sharply increase the risk of collision. Our model proposes uncertainty-based adaptive planning to detect the risk of an impending collision and trigger an appropriate re-planning of the trajectory.}
    \label{fig:main_fig}
\end{figure*}

The rest of the paper is structured as follows. In Section \ref{sec:related_work} we review related work in this field, emphasizing collision avoidance, planning using generative models and the estimation of uncertainty in neural networks. In Section \ref{sec:methods} we formulate the problem of collision avoidance using offline RL and uncertainty estimation through deep ensembles. In Section \ref{sec:results_and_discussion} we demonstrate the effectiveness of our collision avoidance algorithm and present our experimental results on the \texttt{highway-env} environment while discussing our key findings. Finally, in Section \ref{sec:conclusion} we summarize our results, discuss the limitations of our model and outline future research directions.

\section{Related Work}\label{sec:related_work}
\subsection{Collision Avoidance in Motion Planning}

Collision avoidance is of fundamental importance in robotics, particularly in environments with dynamically moving obstacles. Factors such as lack of communication between multiple users in an environment and lack of a central planner make avoidance of collision with dynamic obstacles challenging. Traditional, rule-based approaches include methods such as dynamic window approach (DWA) \cite{fox_1997_dynamic} and timed elastic band (TEB) \cite{rosmann_2015_timed}, which are, however, limited to static obstacles. The Velocity Obstacle (VO) series method \cite{fiorini_1998_motion, van_2008_reciprocal, van_2011_reciprocal, snape_2011_hybrid} was proposed to tackle the problem of dynamic obstacle avoidance. However, these methods involve significant computational overhead due to their reliance on conventional rules defined by complex conditions and equations.

More recent methods involve learning to avoid collisions by utilizing deep neural networks in reinforcement learning to simplify complex rules and conditions. Principal collision avoidance methods in this category include CADRL (collision avoidance with deep reinforcement learning) \cite{chen_2017_decentralized, everett_2021_collision} and MRCA (Multi-robot collision avoidance) \cite{long_2018_towards}. While both CADRL and MRCA rely on the prediction of future states of obstacles to avoid a collision, MRCA requires continuous communication between the multiple agents. On the other hand, CADRL utilizes reinforcement learning to enable an agent to learn collision avoidance behaviors from simulated experiences without any communication between the agents. Similar other approaches to collision avoidance based on deep reinforcement learning include \cite{qiao_2008_application, zhang_2016_robot, cao_2022_multi, chen_2022_deep, xue_2019_deep, choi_2021_reinforcement}.

\subsection{Planning using Generative Models}
Recently, planning using generative models such as transformers and diffusion models has gained significant attention in the context of offline RL, where the goal is to learn optimal policies from pre-collected datasets without further interaction with the environment. Generative models are used to generate trajectories of future states conditioned on the current state, enabling robust planning in the absence of real-time feedback. This approach is especially effective in tackling problems where exploration is costly or risky, such as in robotics and autonomous driving, by improving the agent's ability to generalize from limited offline data \cite{chen_2024_deep}.

One of the representative works in this area is Decision Transformer \cite{chen_2021_decision} (DT), which casts the sequential decision-making problem into a sequence-modeling task, and solves it using a Transformer \cite{vaswani_2017_attention}. DT utilizes a decoder-only GPT-style transformer \cite{radford_2019_language}, leveraging its self-attention mechanism to model trajectories as sequences of states, actions, and returns. Given an offline dataset of trajectories $\left\{\tau = \left(s_0, a_0, R_0, \dots s_T, a_T, R_T\right)\right\}$, where $R_t = \sum_{i=t}^Tr_i$ denotes the cumulative future returns, DT is trained to predict the next action based on the previous $k+1$ transitions \cite{chen_2024_deep}:
\begin{equation*}
    \operatorname{min}_\pi \mathbb{E}_\tau\left[ \sum_{t=0}^T -\operatorname{log}\pi\left(a_t | \tau_{t-k:T}\right) \right]
\end{equation*}
DT simplifies offline RL by eliminating the need to fit Q-value networks through dynamic programming or computing policy gradients, and instead utilizes supervised sequence modeling. This has eventually evolved into a large class of transformer-based algorithms, and has been collectively referred to as Transformer-based RL (TRL) \cite{hu_2024_transforming}.  

More recently, diffusion models, which have achieved significant success in image and video generation, have been effectively applied to trajectory modeling within offline RL. Diffusion models have been notably used as planners, where they are trained to generate a trajectory clip $\tau = \left(e_1, e_2, \dots e_H\right)$, where $H$ is the planning horizon. Possible choices for $e_t$ include $e_t = \left(s_t, a_t\right)$ \cite{janner_2022_planning}, $e_t = s_t$ \cite{ajay_2023_conditional}, and other combinations. A prominent work in this domain is the Decision Diffuser (DD) \cite{ajay_2023_conditional}, which leverages diffusion processes to model the distribution of future trajectories based on past observations. DD operates by progressively refining noisy samples of potential future state sequences, conditioned on the desired outcomes, allowing it to generate trajectories that align with target rewards or goals. Unlike standard generative models, diffusion models excel at modeling complex distributions, making them particularly suited for environments with high variability and uncertainty. Other principal works in this area include Diffuser \cite{janner_2022_planning} and SafeDiffuser \cite{xiao_2024_safediffuser}. 

\subsection{Uncertainty Estimation in Neural Networks}
Uncertainty estimation is an essential component in a wide range of tasks, particularly in those involving decision-making in safety-critical applications such as autonomous driving and robotics. The predictive uncertainty of a neural network consists of two components: epistemic uncertainty, which is the uncertainty associated with a model's knowledge, and aleatoric uncertainty, which is associated with the noise in the data \cite{gawlikowski_2023_survey}. 

Traditionally, uncertainty quantification has been approached through Bayesian inference, where a prior distribution is placed on a network's parameters, and the posterior distribution is computed over the training data. The goal is to compute the posterior distribution $p\left(\theta | \mathcal{D}\right)$ where $\mathcal{D}$ is the dataset, which can then used to compute the uncertainty. However, since exact Bayesian inference is intractable, various approximation techniques have been proposed.

One such method is Monte Carlo Dropout \cite{gal_2016_dropout}, which approximates Bayesian inference by applying dropout both during training and inference. By performing multiple stochastic forward passes, we obtain a sample of network outputs  $\{ \hat{y}_i\}_{i=1}^{N}$, where each pass provides an approximation of the posterior, and the mean and variance of these outputs are used to estimate predictive uncertainty

Deep ensembles \cite{lakshminarayanan_2017_simple} offer a more straightforward yet highly effective alternative for uncertainty estimation. In this approach, multiple independent models $\{f_{\theta_k} \}_{k=1}^{K}$  are trained from different random initializations, each representing a different mode of the posterior distribution. The ensemble prediction is formed by averaging the predictions from each model, while the variance across the model outputs serves as an estimate of the model’s uncertainty. Deep ensembles are highly scalable, as they don't need complex posterior approximations while providing strong uncertainty estimates. Deep ensembles capture both the epistemic and aleatoric uncertainties and do so without requiring any changes to the prediction network architecture.

\section{Methods}\label{sec:methods}
\subsection{Problem Description}
We consider a Markov Decision Process (MDP) with state space $\mathcal{S}$ and action space $\mathcal{A}$.
The dynamics of the MDP are governed by a stochastic transition function $\mathcal{T} : \mathcal S \times \mathcal A \mapsto \mathcal S$. 

The reward function is denoted by $\mathcal{R}: \mathcal{S}\times\mathcal{A} \rightarrow \mathbb{R}$. A trajectory $\tau$ consists of a sequence of states, actions and rewards: 
\begin{equation}
    \tau := \left\{\left(s, a, r\right)\right\}_{t=t_1}^{t_N} \text{ for } t \in \{t_1, t_2 \dots t_N\} 
\end{equation}
where $(s, a, r)_t \in \mathcal{S}\times\mathcal{A}\times\mathbb{R}$. The return $R(\tau)$ is defined as the sum of rewards over all time steps within a trajectory: $R(\tau) = \sum_tr_t$. The goal is to learn an optimal policy $\pi^*$ which maps the environment's state to an agent's action $\pi: \mathcal{S} \rightarrow \mathcal{A}$ such that it maximizes the expected return over all trajectories, i.e. 
$\pi^* = \operatorname{argmax}_{\tau \sim \pi}\mathbb{E} \left[ R(\tau) \right]
$.

Conventional approaches to obtaining an optimal policy include Q-learning, Deep Q-Networks (DQN) and Policy Gradient Methods under the online reinforcement learning paradigm, where the agent continuously interacts with the environment to learn a policy.

Recent advancements in reinforcement learning have enabled offline RL, where an agent learns an optimal policy \( \pi^* \) from a static dataset \( \mathcal{D} \) consisting of pre-collected trajectories \( \tau = \{(s_t, a_t, r_t)\}_{t=1}^T \), rather than through continuous interaction with the environment. This dataset typically contains sub-optimal trajectories generated by various policies, with each trajectory labeled with corresponding rewards. The goal is to infer a return-maximizing policy \( \pi^*(a|s) \) using the data in \( \mathcal{D} \). Offline RL offers significant advantages, particularly in environments where exploration is risky or costly (e.g., autonomous driving), as the agent learns exclusively from pre-existing data without interacting with the real world. Additionally, offline RL allows the integration of large, diverse datasets from multiple sources, increasing the robustness of the learned policy, while mitigating the dangers of unsafe exploration.

% Recent advancements have enabled offline reinforcement learning, in which a dataset consisting of many sub-optimal, reward-labeled trajectories is utilized to enable the learning of an optimal policy without continuous interaction with the environment. Offline RL is advantageous in several ways: in environments where exploration can be potentially risky, such as in autonomous driving, learning from an offline dataset ensures safety. Further, offline RL allows for the use of extensive datasets, potentially collated from multiple different sources, to learn a policy. 

\subsection{Diffusion Models for Offline Reinforcement Learning}\label{subsec:diffusion_models_for_offline_reinforcement_learning}
Diffusion models have emerged as a powerful tool for modeling complex distributions in modalities such as images and videos. In the context of offline RL, the task of learning an optimal policy from a static dataset can be framed as a sequence modeling task, where trajectories $\tau = \{e_t\}_{t=1}^T$ are treated as sequences. Here $e_t$ could be state-action pairs $(s_t, a_t)$ \cite{janner_2022_planning}, just $s_t$ \cite{ajay_2023_conditional} or other possible options. Following \cite{ajay_2023_conditional}, we choose to use $e_t = s_t$ and exclude the actions from the sequences, in consideration of the fact that actions sequences are usually high-frequency, making them harder to model. We use a separate network, called the inverse dynamics action model (described in Section \ref{subsec:inverse_dynamics_action_model}) to map the state sequence back to actions: $f_\theta(s_t, s_{t+1}) = a_t$. 

We use a Denoising Diffusion Probabilistic Model (DDPM) \cite{ho_2020_denoising} to model the state sequences. DDPM consists of a forward process in which noise is progressively added to an input sample until it becomes pure noise, and a reverse process in which the model learns to reverse the noising process through a kernel to progressively generate trajectories from pure noise. Given an input sample $\mathbf{x}_0$ from a distribution $p_{\text{data}}(\mathbf{x}_0)$, the forward noising process produces a sequence of noisy vectors $\mathbf{x}_0$, $\mathbf{x}_1$ $\dots$ $\mathbf{x}_K$ with the transition kernel:
\begin{equation*}
    q\left(\mathbf{x}_{k+1} | \mathbf{x}_{k}\right) = \mathcal{N}\left(\mathbf{x}_{k+1};\sqrt{\alpha_K}\mathbf{x}_k; (1-\alpha_k)I\right)
\end{equation*}
where $\alpha_k$ follows a noise schedule. As the number of diffusion steps $k \rightarrow \infty$, the final state distribution $q_K$ converges in distribution to a standard normal distribution $\mathcal{N}(0, I)$. The reverse process is accomplished using successive applications of a learnable kernel:
\begin{equation*}
    p_\theta\left(\mathbf{x}_{k_1} | \mathbf{x}_k\right) = \mathcal{N}\left(\mathbf{x}_{k-1}| \mu_\theta \left(\mathbf{x}_k, k\right), \Sigma_k\right)
\end{equation*}
The kernel is trained to match the intermediate noisy vectors in the forward process and the loss function takes the form:
\begin{equation*}
    \mathbb{E}_{k\sim\mathcal{U}[1, K], \mathbf{x}_0\sim p_{\text{data}}(\mathbf{x}_0), \epsilon\sim\mathcal{N}(0, I)}\left[\left\lVert\epsilon - \epsilon_\theta(\mathbf{x}_k, k)\right\rVert^2\right]
\end{equation*}
where $\mathcal{U}[1, K]$ is a discrete uniform distribution over $\{1, 2, \dots K\}$ and $\epsilon_\theta$ is a deep neural network which predicts the noise $\epsilon$ from $\mathbf{x}_k$ and $k$.
\subsection{Inverse Dynamics Action Model}\label{subsec:inverse_dynamics_action_model}
The diffusion model generates a state sequence $\hat{s}_{1:T} \sim p_\theta(s_{1:T} | s_0)$ given an initial state $s_0$. In order to learn the policy, we must learn the action sequences that enable the transitions in this state sequence. We accomplish this by using an inverse dynamics model, which predicts the action that effects the transition between a pair of consecutive states: $f_\theta(s_t, s_{t+1}) = a_t$. The action model is trained using the offline dataset to learn the transition kernel of the environment with cross entropy loss as the criteria for optimization.

\subsection{Uncertainty-based Obstacle Detection using Deep Ensemble}
In a dynamically changing environment with moving obstacles, the risk of collision between the agent and obstacles is significantly higher than in a static or slowly varying environment. Therefore, long-horizon planning described in \ref{subsec:diffusion_models_for_offline_reinforcement_learning} is not well suited for such dynamically changing environments often encountered in real life. On the other extreme, generating a trajectory at each time step, conditioned on the current state, would enable dynamic planning and lower the risk of collision, but it is computationally expensive and highly redundant during the times when obstacles do not pose a real risk. Therefore, we propose an adaptive strategy that triggers re-planning only when we detect that the uncertainty of our planning process is above a set threshold. This method is based on the idea that increased uncertainty indicates a stronger need to reassess the plan using the generative model, ensuring that future decisions are informed by the latest environmental observations, aiding in collision avoidance.

Following \cite{jutras_2024_adaptive}, we quantify this uncertainty using a Deep Ensemble of action models. The predictive uncertainty of deep ensembles captures both aleatoric (arising from the noise in the environment's transitions) and epistemic uncertainties (arising from lack of model's knowledge). In the case of discrete actions, an action model outputs the probability of each action:
\begin{equation*}
    f_\theta(s_t, s_{t+1}) = \left[p(a_t^0), p(a_t^1), \dots, p(a_t^{K-1})\right]
\end{equation*}
where $K$ is the total number of possible actions.
This vector is obtained by applying softmax operation on the final linear layer of the action model. The deep ensemble consists of identical action models, each trained on the same dataset, but initialized with random parameters to increase the variation in their predictions. The predicted action probabilities of the ensemble are taken to be the mean predictions of all models in the ensemble:
\begin{equation*}
    \left[p(a_t^0), p(a_t^1) \dots p(a_t^{K-1})\right] = \frac{1}{M}\sum_{m=1}^M f_{\theta_m}(s_t, s_{t+1}).
\end{equation*}
The total predictive uncertainty of the action model is given by the total entropy in the predictions:
\begin{equation}
    u_t = -\sum_{k=0}^{K-1}p(a_t^k)\log p(a_t^k).
\end{equation}
The taken action is 
\begin{equation}
    a_t = \operatorname{argmax}\left[p(a_t^0), p(a_t^1) \dots p(a_t^{K-1})\right]
\end{equation}
In the course of a trajectory $\tau = \left(s_1, s_2 \dots s_T\right)$, we trigger re-planning using the diffusion model at time $t$ when $u_t > \epsilon$, where $\epsilon$ is a tunable set threshold. Thus, this approach provides us a means to adaptively balance long-horizon planning and collision safety, and tune this balance using a single parameter, $\epsilon$.

\begin{algorithm}
\caption{Adaptive Collision Avoidance with Uncertainty Estimation}\label{alg:adaptive_planning}
\begin{algorithmic}[1]
\Require Generative state model $p_\theta(s)$, action model ensemble $E = \{f_{\phi_1}, \dots, f_{\phi_M}\}$, threshold $\epsilon$
\State $t \gets 0$, observe initial state $s_0$
\While{episode not done}
    \State Sample trajectory of future states $\hat{s} \sim p_\theta(s|s_t)$
    \State Predict $a_t$, $u_t$ using ensemble $E$
    \State Execute $a_t$, increment $t$, observe new $s_t$
    \State Set $\hat{s}_1 \gets s_t$
    \State Predict next $a_t$, $u_t$ using ensemble $E$
    \State $i \gets 1$
    \While{$i < H - 1$ and $u_t < \epsilon$}
        \State Execute $a_t$, increment $t$, observe new $s_t$
        \State Update $\hat{s}_i \gets s_t$
        \State Predict next $a_t$, $u_t$ from $E$
        \State $i \gets i + 1$
    \EndWhile
\EndWhile
\end{algorithmic}
\end{algorithm}

\section{Results and Discussion}\label{sec:results_and_discussion}
We perform experiments on the \texttt{highway-env} environment \cite{highway-env}, where an agent vehicle is tasked with navigating through a multi-lane road, which is shared by other vehicles, which act as moving obstacles for the agent. We choose a 4-lane highway environment, with total vehicle count set to 200 and maximum episode length of 100. The offline dataset for this environment is generated by training a Proximal Policy Optimization (PPO) algorithm \cite{schulman_2017_proximal} implemented by Stable Baselines3 \cite{raffin_2021_stable}. The trained model is then used to sample trajectories with a cumulative length of $N \approx 10^6$ steps, which is then used to train the diffusion model and the action model ensemble, as described in Section \ref{sec:methods}.

\begin{table*}
  \centering
  
  \begin{tabular}{lccccc}
    \toprule
     & \multicolumn{4}{c}{Performance} & \multicolumn{1}{c}{Computational Efficiency} \\
    \cmidrule(r){2-5} \cmidrule(r){6-6}
    \textbf{Model}     & \textbf{Mean Trajectory Length}  & \textbf{Num. Collisions} & \textbf{Mean Reward} & \textbf{Mean High-Speed Reward} & \textbf{Saved NFE (\%)} \\
    \midrule
    \textbf{Adaptive (ours)} & 83.2 ($\pm 30.6$)  & 3 & 62.8 ($\pm 23.4$) & 0.17 & 86.7    \\
    DD (no replan) & 73.3 ($\pm 40.2$)  & 4 & 55.7 ($\pm 30.6$) & 0.14 & 97.7  \\
    DD (continuous replan) & 96.6 ($\pm7.2$)  & 2 & 71.25 ($\pm 6.28$) & 0.19 & 0.0  \\
    \bottomrule
  \end{tabular}
  % \vspace{10pt}
  \caption{This table presents the performance metrics and computational efficiency of our adaptive model, alongside DecisionDiffuser in two modes: long-horizon (no re-planning) and continuous re-planning. Metrics shown are averages from 10 evaluations with random initialization. Our adaptive strategy strikes a balance between the two extremes of DecisionDiffuser, achieving a trade-off between collision safety and computational cost. }
  \label{table:results1}
\end{table*}

\subsection{Mean Trajectory Length}

We first evaluate the performance of our adaptive planning strategy using the metric of mean trajectory length, which reflects the agent’s ability to avoid collisions and navigate the environment safely, over a maximum of 100 simulation steps. We choose the uncertainty threshold value to be $\epsilon=0.1$.

\begin{figure}
    \centering
    \includegraphics[width=0.8\columnwidth]{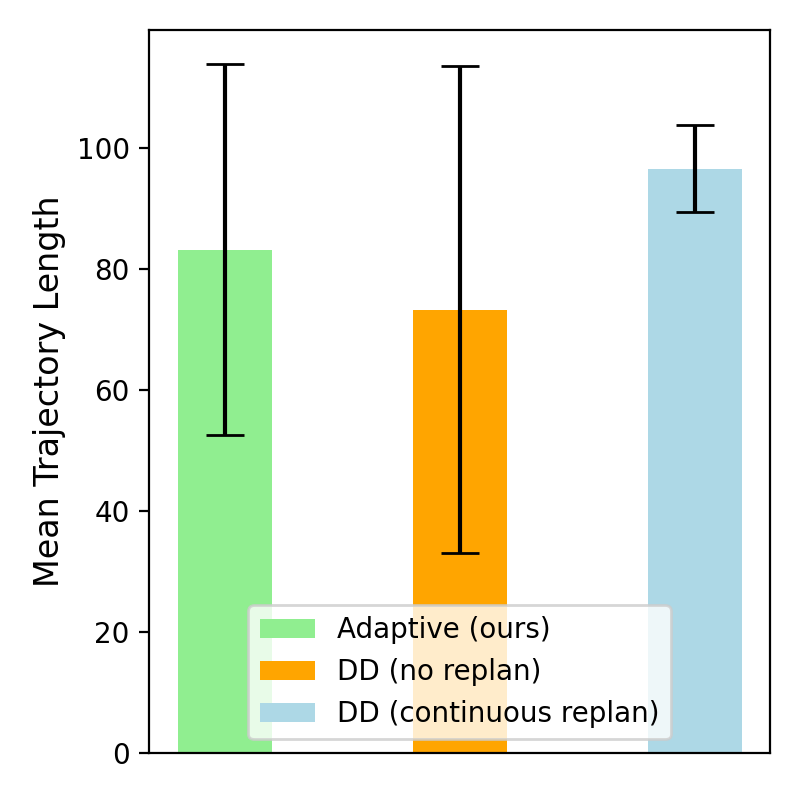}
    \caption{Mean trajectory length (out of a maximum of 100 steps) is shown for three approaches: (a) Adaptive replanning based on uncertainty estimates (\textbf{ours}) with $\epsilon=0.1$, (b) Decision Diffuser (DD) long-horizon, i.e. with no replanning and (c) DD with continuous replanning at every time step. Results are shown for $n=10$ episodes with random initialization.}
    \label{fig:mean_traj_len}
\end{figure}

Fig.~\ref{fig:mean_traj_len} shows that the mean trajectory length given by our adaptive re-planning approach (83.2 steps) is higher than that of long-horizon planning (73.3 steps), an increase of 13.5\%. This result indicates a reduction in collision rates, as longer trajectories suggest that the agent successfully avoids collisions and completes its tasks without needing to terminate early due to obstacle interference. This demonstrates the effectiveness of our uncertainty-driven re-planning mechanism, which dynamically adjusts the agent’s actions only when necessary, optimizing both safety and computational efficiency.

In contrast, the mean trajectory length for the adaptive re-planning strategy is lower than that of re-planning at every step (96.6). This is expected, as re-planning at each time step ensures the highest level of responsiveness to dynamic changes in the environment, although at a much higher computational cost than our adaptive approach. The key advantage of adaptive re-planning lies in maintaining approximately the same level of collision safety as step-wise re-planning while significantly reducing computational overhead.

\subsection{Collision Rate}
We next analyze the collision rate, defined as the percentage of episodes in which the agent experiences a collision during its trajectory. Over 10 evaluation episodes starting from a random state, we find that our approach yields a reduction in the collisions from 4 (for the case of no re-planning) to 3 (Table \ref{table:results1}). On the other hand, the lower collision of the constant re-planning approach (2) demonstrates its ability to continuously adapt to dynamic obstacles, but this comes at the expense of computational efficiency which is quantified by the percentage of saved network function evaluations (NFE). As seen in Table~\ref{table:results1}, our adaptive strategy saves about 86.7\% of NFEs, indicating significant savings in computational resources.

\begin{figure}
    \centering
    \includegraphics[width=0.9\columnwidth]{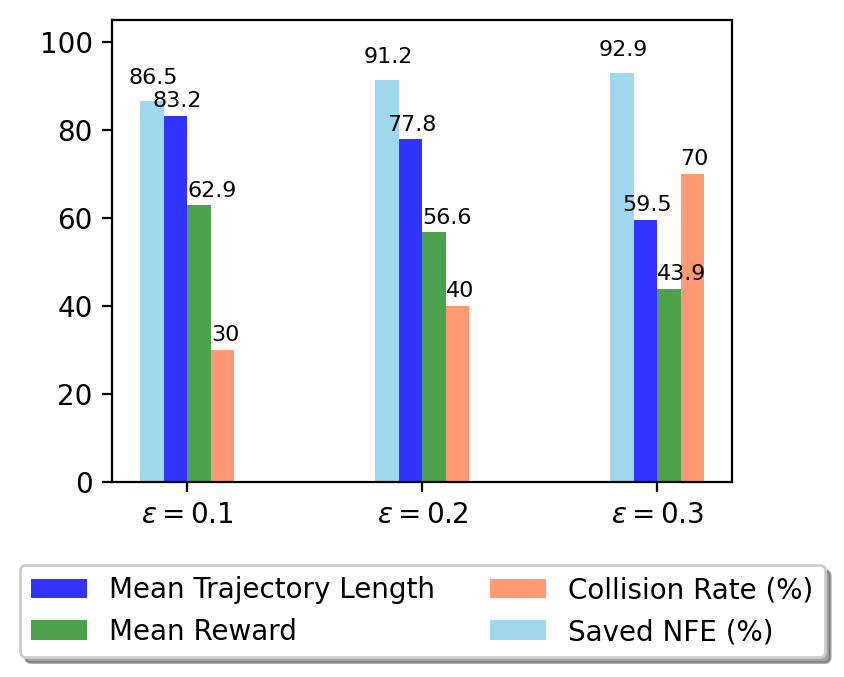}
    \caption{This plot illustrates the impact of varying uncertainty threshold value $\epsilon$ on key performance metrics: mean trajectory length, mean reward, and collision rate. As the threshold value increases, a decline in both mean trajectory length and reward is observed, accompanied by a corresponding rise in the collision rate, highlighting the impact of lower rate of adaptive re-planning when the threshold is higher.}
    \label{fig:epsilon_sweep}
\end{figure}

% \subsection{Tunability of the Re-planning Threshold}

% To evaluate the flexibility and tunability of the proposed adaptive re-planning strategy, we examined the effect of adjusting the re-planning uncertainty threshold on both the collision rate and the frequency of re-planning events. Figure 2 presents a plot of collision rates versus threshold values. As the threshold increases, we observe a corresponding increase in collision rates, as expected, since higher thresholds delay the re-planning process, reducing the system’s responsiveness to dynamically changing obstacles.

% At the same time, the number of re-planning events decreases as the threshold increases (Figure 3). This trade-off highlights the tunability of the system: lower thresholds prioritize safety by triggering more frequent re-planning, while higher thresholds reduce computational overhead at the cost of a slight increase in collision risk. This tunability allows the system to be adjusted based on the specific requirements of the task, whether that be minimizing collisions or reducing computational load.

\subsection{Impact of uncertainty threshold values}
Selecting a high uncertainty threshold reflects a reluctance to re-plan unless the model exhibits high uncertainty regarding its actions. This is expected to negatively impact collision safety, as the system becomes less responsive to dynamic changes. At the same time, we also expect a reduction in the frequency of re-planning, resulting in savings in NFE (Number of Function Evaluations). This trade-off between safety and computational efficiency is consistently observed in our experiments, as seen in Fig.~\ref{fig:epsilon_sweep}

In summary, the adaptive re-planning strategy demonstrates clear advantages over long-horizon diffusion planning by reducing collision rates and offering a tunable trade-off between safety and efficiency. The ability to adjust the re-planning threshold provides further flexibility, enabling the system to balance collision avoidance with computational overhead according to task demands.

\section{Conclusion}\label{sec:conclusion}

This work presents a novel approach for enhancing collision avoidance in dynamically changing environments by leveraging uncertainty estimates from a deep ensemble of inverse dynamics action models alongside a diffusion model for trajectory planning. Our method focuses on improving the safety and robustness of trajectory generation in scenarios where obstacles are constantly moving and the environment is highly unpredictable. By using a diffusion model to generate long-horizon trajectories and selectively re-planning based on uncertainty estimates, we strike a balance between collision safety and computational cost.

The results show that the proposed approach leads to longer mean trajectory lengths, indicating successful collision avoidance without sacrificing computational efficiency. Importantly, the tunability of the re-planning threshold allows for a fine balance between minimizing collisions and managing computational load, making it adaptable to a variety of real-world scenarios where responsiveness and safety are important.

Our approach offers a promising solution for applications that require real-time decision-making in dynamic environments, such as autonomous driving, robotics, and drones. By focusing on improving collision avoidance through uncertainty-aware planning, our work contributes to the broader effort of making autonomous systems safer and more reliable in the presence of environmental uncertainty. 
% \addtolength{\textheight}{-8cm}  

% \section*{Appendix}

\section*{Acknowledgement}
This material is based upon work supported in part by the DEVCOM Army Research Laboratory under cooperative agreement W911NF2020221

\bibliographystyle{IEEEtran}
\bibliography{IEEEabrv, references}

\end{document}